\documentclass{article}

\usepackage[final]{corl_2018} 
\usepackage{wrapfig}

\usepackage{subcaption}
\usepackage{graphicx}

\usepackage{amsmath}
\usepackage{amsfonts}
\usepackage{mathtools}

\usepackage[utf8]{inputenc} 
\usepackage[T1]{fontenc}    
\usepackage{hyperref}       
\usepackage{url}            
\usepackage{booktabs}       
\usepackage{amsfonts}       
\usepackage{nicefrac}       
\usepackage{microtype}      
\usepackage[svgnames]{xcolor}
\usepackage{authblk}

\usepackage{makecell}

\title{Robustness via Retrying: Closed-Loop Robotic Manipulation with Self-Supervised Learning}

\author[1]{Frederik Ebert}
\author[1]{Sudeep Dasari}
\author[1]{Alex X. Lee}
\author[1]{Sergey Levine}
\author[1]{Chelsea Finn}

\affil[1]{\footnotesize Department of Electrical Engineering and Computer Sciences, UC Berkeley, United States}

\affil[ ]{\texttt{\{febert,sdasari,alexlee\_gk,svlevine,cbfinn\}@berkeley.edu}}

\begin{document}

\maketitle

\begin{abstract}
Prediction is an appealing objective for self-supervised learning of behavioral skills, particularly for autonomous robots. However, effectively utilizing predictive models for control, especially with raw image inputs, poses a number of major challenges. How should the predictions be used? What happens when they are inaccurate? In this paper, we tackle these questions by proposing a method for learning robotic skills from raw image observations, using only autonomously collected experience. We show that even an imperfect model can complete complex tasks if it can continuously retry, but this requires the model to not lose track of the objective (e.g., the object of interest). To enable a robot to continuously retry a task, we devise a self-supervised algorithm for learning image registration, which can keep track of objects of interest for the duration of the trial. We demonstrate that this idea can be combined with a video-prediction based controller to enable complex behaviors to be learned from scratch using only raw visual inputs, including grasping, repositioning objects, and non-prehensile manipulation. Our real-world experiments demonstrate that a model trained with 160 robot hours of autonomously collected, unlabeled data is able to successfully perform complex manipulation tasks with a wide range of objects not seen during training.

\end{abstract}

\section{Introduction}

\begin{wrapfigure}{r}{.4\columnwidth}
\vspace{-5mm}
\centering
\includegraphics[width=0.4\columnwidth,trim={3.2mm 0 0 0},clip]{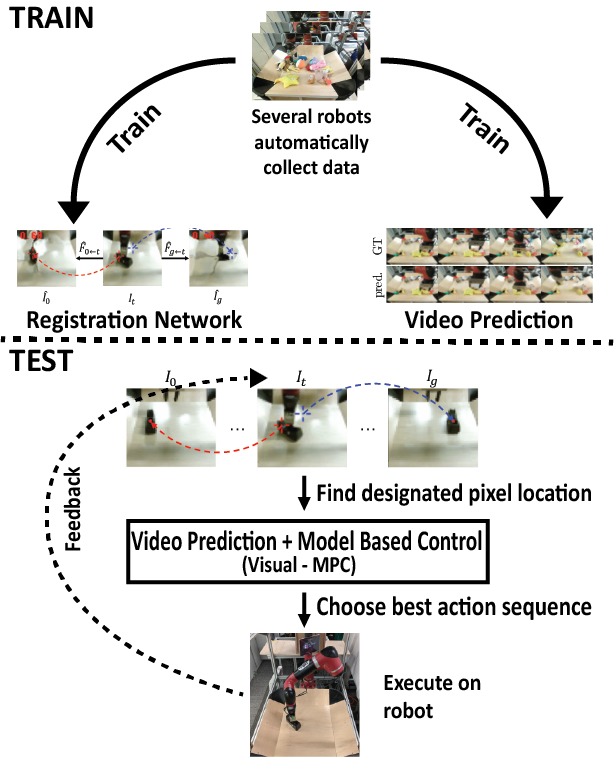}
\caption{\small{Closed-loop visual MPC: Autonomously collected experience is used to train a video prediction model, as well as an image-to-image registration model, which enables the robot to keep track of its goal.}}
\label{fig:overview}
\vspace{-0.3in}
\end{wrapfigure}

Humans have the ability to learn complex skills, such as manipulating objects, through millions of interactions with their environment during their lifetime.
These interactions enable us to acquire a general understanding of the physical world and, notably, do not require significant supervision beyond observation of one's own actions and their consequences. Hence, self-supervised learning through prediction is an appealing direction of research as it enables intelligent systems to leverage and learn from massive amounts of unlabeled raw data to autonomously acquire general skills. Yet, self-supervised learning systems using predictive models of sensory inputs present a number of challenges: planning needs to account for imperfections in the predictive model and the robot needs a grounded mechanism for evaluating predicted futures.
How can we enable systems to plan to perform complex tasks from raw sensory observations, even when the predictions are not always accurate?

Prior work on self-supervised robot learning has enabled robots to learn rudimentary, short-term manipulation skills such as grasping~\cite{lerrel,google_handeye}, singulation~\cite{princeton_pushgrasp}, pushing~\cite{foresight,sna}, and poking~\cite{pulkit}. The question that we are concerned with in this work is: can self-supervised predictive models of raw visual observations be used to perform more complex and realistic tasks, especially tasks that are temporally extended? 
This might seem challenging due to the difficulty of long-term prediction of image observations. However, if we can enable a robot to continuously correct its behavior during execution and retry in the event of failure or misprediction, even a simple planning algorithm with short-term predictive ability can succeed at the task. This amounts to applying the well-known principle of model-predictive control (MPC), which is typically used with analytical models and low-dimensional state, to control via video prediction. Hence, we call this method visual MPC.

We propose a cost function for vision-based control based on image-to-image registration, which we demonstrate can itself be learned without any supervision using the same exact dataset as the one used to train the predictive model. The key element that enables our method to perform long-horizon tasks is that, using our cost function, the robot can always evaluate the distance to the goal, allowing it to continuously retry, so that even flawed predictions allow for an eventual successful execution.

We demonstrate our method on the task of maneuvering unknown objects in a table-top setting using a robot manipulator. To autonomously learn to perform manipulation skills accurately, tasks need to be specified in way that allows for precision and retrying. We specify a goal by providing an image of the desired configuration along with user-annotated positions of the objects or points of interest\footnote{This also allows the user to specify distractor objects that can be ignored in the goal image.}. This provides a straight-forward and grounded mechanism for a user to specify a goal in the observation space of the robot. Note, however, that these commands are not provided during training, and the model has no prior knowledge of objects or goals.
Building on prior methods that use self-supervised predictive models for control~\cite{foresight,sna,se3_control}, we develop a method that can plan actions with a video prediction model to achieve the desired state specified in the goal image.

The main contribution of this work is a method for computing the planning cost for visual MPC based on image-to-image registration, using a learned registration model to put the current observation in correspondence with both the initial and goal images. See \autoref{fig:overview} for an illustration.
This allows closed-loop control, where the robot persistently attempts the task until completion. In contrast to the short-horizon pushing skills demonstrated in prior work~\cite{foresight,sna}, we show that our video prediction model can be used to perform longer-horizon manipulations and succeed more consistently. We also demonstrate that this approach can enable a robot to autonomously choose between prehensile and non-prehensile manipulation without any supervision, choosing when to push or pick up objects to relocate them to achieve the goal. Finally, we extend this method to visual MPC with multiple camera videos, showing that task goals can be specified in 3D via prediction from stereo observations.

\vspace{-0.1cm}
\section{Related Work}
\vspace{-0.1cm}

While object relocation with both non-prehensile \cite{hermans2013learning,salganicoff1993vision} and prehensile \cite{goldfeder2009data} manipulation strategies has been explored extensively in prior work, much of this research has focused on hand-designed models and feature representations, which require extensive engineering and make strong assumptions about the environment and task.
Video-prediction based manipulation is more general, because it does not require any human intervention at training time, and learns about the physics of object interaction directly from data. A more subtle point is that direct prediction of camera images makes no assumptions about the representation of the world state, such as the number of objects.
Many self-supervised robot learning methods have focused on predicting the outcome of a particular event, such as grasp success~\cite{lerrel,google_handeye,princeton_pushgrasp} or crashing~\cite{crashing,greg_kahn_uncertainty}. Consequently, these approaches can only recover a single policy that optimizes the probability of the predicted event. Instead, we aim to acquire a model that can be reused for multiple goals and behaviors. To do so, we build upon prior works that learn to predict the sequence of future observations, which can be used to optimize with respect to a variety of goals~\cite{foresight,sna,se3_control}. Unlike~\cite{se3_control}, we demonstrate complex object manipulations with previously-unseen objects from RGB video. In contrast to prior self-supervised visual planning works~\cite{foresight,sna}, we can perform substantially longer tasks, by using image registration with respect to a goal image.

Goal observations have been previously used for specifying a reward function for robot learning systems~\cite{jagersand1995visual,deguchi1999image,e2c,dsae}. Unlike these methods, we use a learned registration to the goal image for measuring distance to the goal rather than distance in a learned feature space. Distances in unconstrained feature spaces can be arbitrary, while registration inherently measures how pixels should move and can therefore provide a calibrated distance metric with respect to the goal. 

A related problem is visual servoing, where visual feedback is used to reach a goal configuration~\cite{hutchinson1996tutorial,kragic2002survey,desouza2002survey}.
Traditional approaches aim to minimize distances of feature points~\cite{feddema1989vision,espiau1992servo,wilson1996relative}, or pixel intensities~\cite{caron2013photometric}. Learning and convolutional networks have also been incorporated into visual servoing~\cite{saxena2017servoing,bateux2018servoing,lee2017servoing,google_handeye}. Unlike servoing approaches that use reactive control laws, we use multi-step predictive models to achieve temporally extended goals, while still using continuous visual feedback for retrying at every step. Further, our method performs non-prehensile manipulation, while visual servoing typically assumes fully actuated control, often with a known Jacobian.

Model-predictive control (MPC)~\cite{camacho2013model} has proven successful in a variety of robotic tasks~\cite{shim2003decentralized,allibert2010predictive,howard2010receding,williams2017information,deep_mpc}.
MPC involves planning with a short horizon and re-planning as the execution progresses, providing the foundation of persistent retrying and the ability to overcome inaccuracies in the predictive model. However, as we later show, maintaining an accurate model of the \emph{cost} used for planning throughout an episode is critical, and has prevented prior work on visual foresight~\cite{foresight,sna} from moving beyond short-term tasks.
Our primary contributions is a grounded mechanism for evaluating the planning cost of visual predictions, allowing persistent re-planning with video prediction models.

It is possible to use off-the-shelf trackers~\cite{lucas1981iterative,brox2004high,babenko2009visual,mei2009robust} to mitigate this issue. However, these trackers usually only have a limited capability to adapt to the domain they are applied to, and can lose track during occlusions. A key advantage of our learned registration approach, inspired by \cite{meister2017unflow}, is that we can train our registration model using data collected autonomously by the robot. Since this model is completely self-supervised, it continues to improve as the robot collects more data.

\vspace{-0.1cm}
\section{Preliminaries}
\label{sec:prelim}
\vspace{-0.2cm}

Our visual MPC problem formulation follows the problem statement outlined in prior work~\cite{foresight}. In this setting, an action-conditioned video prediction model $g$, typically represented by a deep neural network, is used to predict future camera observations $\hat{I}_{1:T} \in \mathbb{R}^{T \times H\times W \times 3}$, conditioned on a sequence of candidate actions $a_{1:T}$, where the prediction horizon is $T$. This can be written as $\hat{I}_{1:T} = g(a_{1:T}, I_0)$, where $I_0$ is the frame from the current time-step. An optimization-based planner is  used to select the action sequence that results in an outcome that accomplishes a user-specified goal. This type of vision-based control is highly general, since it reasons over raw pixel observations without the requirement for a fixed-size state space, and has been demonstrated to generalize effectively to non-prehensile manipulation of previously unseen objects~\cite{foresight,sna}.

Visual MPC assumes the task can be defined in terms of pixel motion. In the initial image $I_0$ we define $n$ source pixel locations denoted by the coordinates $d_{0,i} \in \mathbb{N}^2$ (for $i \in [0,..n]$) and the analogous for the goal image $I_g$ denoted by $d_{g,i} \in \mathbb{N}^2$. Given a goal, visual MPC plans for a sequence of actions $a_{1:T}$
to move the pixel at $d_{0,i}$ to $d_{g,i}$. If this pixel lies on top of an object, this corresponds to moving that entire object to a goal position. Note that this problem formulation resembles visual servoing, but it is considerably more complex, since moving the object at $d_0$ might require complex non-prehensile or prehensile manipulation and long-horizon reasoning.
The planning problem is formulated as the minimization of a cost function $c$, which in accordance with prior work \cite{sna}, measures the distance between the predicted pixel positions $\hat{d}_{\tau}$ and the goal position $d_g$ for each pixel $i$:
\begin{align}
c = \sum^n_{i = 1}  \lambda_i c_i && c_i = \sum_{\tau = 1, \dots, T} \mathbb{E}_{\hat{d}_{\tau,i} \sim P_{\tau,i}} \left[\|\hat{d}_{\tau,i} - d_{g,i}\|_2\right]  
\label{eq:cost}
\end{align}
where $c_i\in \mathbb{R}$ are the costs per source pixel, $\lambda_i$ are weighting factors discussed in section \ref{sec:reg} and $P_{\tau,i}$ is the distribution over predicted pixel positions. The advantage of distance-based cost functions is that they are well-shaped and can be optimized efficiently. 

In this paper we use the video prediction model architecture developed by~\cite{savp}, where future images are generated by transforming past images. Starting with a distribution over initial positions of the designated pixel \mbox{$P_{t_0,i}\in\mathbb{R}^{H\times W}, \sum_{H,W} P_{t_0,i} = 1$} at time $t = 0$, the model predicts distributions over its positions $P_{t,i}$ at time $t \in \{ 1, \dots, T \}$ by exploiting the image transformations used to generate future frames. Planning is performed by sampling candidate actions sequences and optimizing using the cross-entropy method (CEM) \cite{cem-rk-13} to achieve the lowest possible cost $c$.

To obtain the best results with imperfect models, the action sequence is replanned at each real-world time step\footnote{We refer to timesteps in the real world as $t$ and to predicted time-steps as $\tau$.} $t \in \{0,...,t_{max}\}$ following the framework of model-predictive control (MPC): at each real-world step $t$, the first action of the best action sequence is executed. 
At the first real-world time step $t=0$, the distribution $P_{\tau=0,i}$ is initialized as 1 at the location of the designated pixel and zero elsewhere. In prior work \cite{sna, foresight}, in subsequent steps ($t > 0$),  the prediction of the previous step is used to initialize $P_{\tau=0,i}$. However this causes accumulating errors, often preventing the model from solving long-term tasks or responding to situations where the outcome of an action was different than expected. In effect, the model loses track of which object was designated in the initial image.

\vspace{-0.1cm}
\section{Retrying by Registration}
\label{sec:reg}
\vspace{-0.2cm}
When using a distance-based cost function for visual MPC it is necessary to update the belief of where the target object currently is, so that the agent can ``keep retrying" indefinitely until the task is accomplished. Prior work on visual MPC lacked this capability. To solve this issue, we propose a method for registering the current image to both the start and goal image, where the designated pixels are known. In this way we can find the locations of the corresponding pixels in the \emph{current image}, allowing us to compute the distances to the goal. Crucially, the registration method we propose is self-supervised, using the same exact data for training the video prediction model and the registration model. This allows both the predictor and registration model to continuously improve as the robot collects more data.

Before further detailing our learned registration system, we discuss a two simple alternative approaches for obtaining a cost-function for video-prediction based control: One na\"{i}ve approach could be to use the pixel-wise error between a \emph{goal image} and the \emph{predicted image}. However there are a number of issues with this approach: first, when objects in the image are far from the position in the goal image (e.g., they do not overlap) there is no gradient signal with respect to changes in the actions. Second, due to the blurry predictions from a video prediction model, the pixel-wise difference between the predictions and the goal image can become meaningless. 

Another approach is to perform a registration between predicted video frames and the goal image, and use the average length of the warping vectors as a cost function for ``closeness" to the goal image. However a major drawback of cost functions based on metrics computed on the complete image is that they naturally emphasize large objects (such as the robot's arm), while small objects only contribute negligible amounts to the costs. As a result, the planner only tries to match the positions of the large objects (the arm), ignoring smaller objects.


\begin{wrapfigure}{r}{.5\columnwidth}
\vspace{-0.25in}
\centering
\includegraphics[width=0.5\columnwidth]{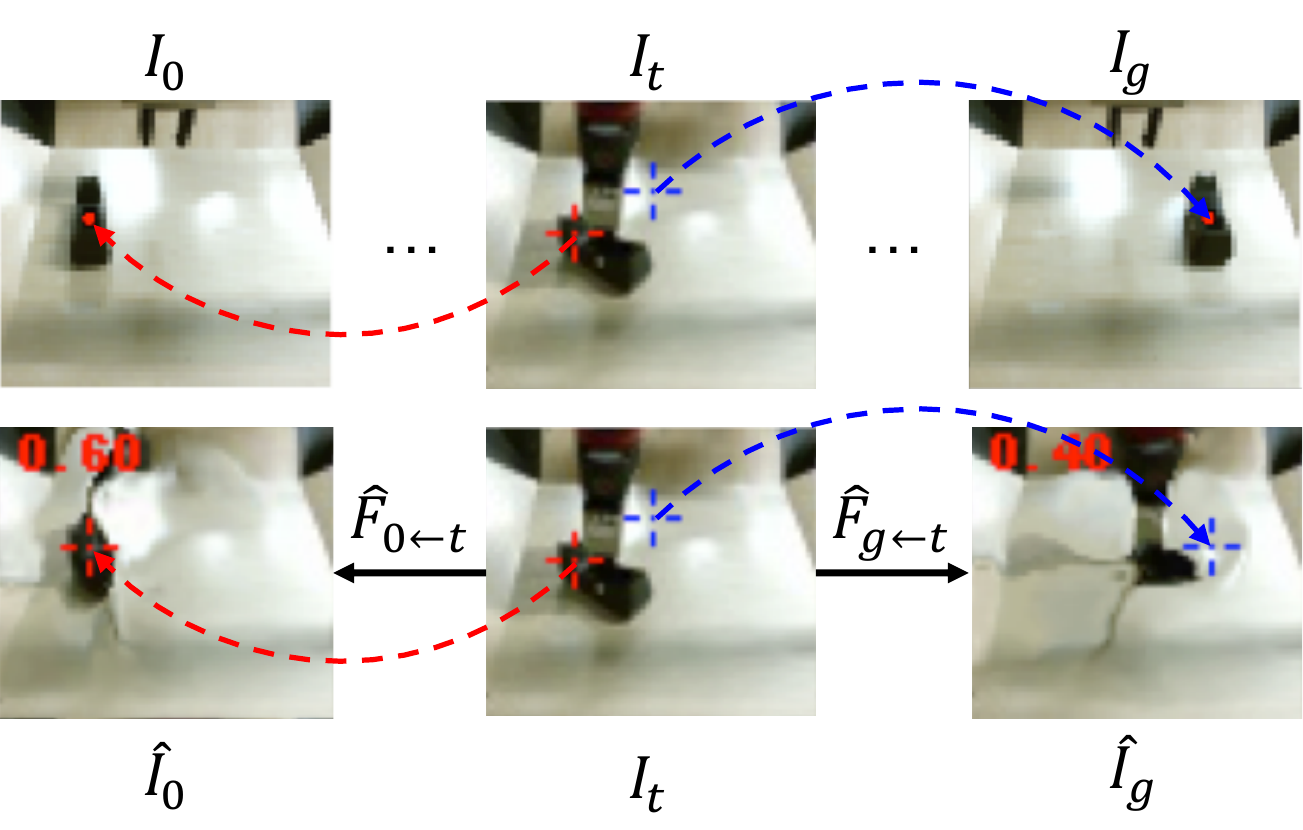}
\caption{\small{Closed loop control is achieved by registering the current image $I_t$ globally to the first frame $I_0$ and the goal image $I_g$. In this example registration to $I_0$ succeeds while registration to $I_g$ fails since the object in $I_g$ is too far away.}
\label{fig:reg_single}
\vspace{-0.2in}
}
\end{wrapfigure}

\subsection{Test time procedure}
\begin{figure}[t!]
    \centering
    \begin{subfigure}[b]{0.25\textwidth}
        \centering
        \includegraphics[width=\textwidth]{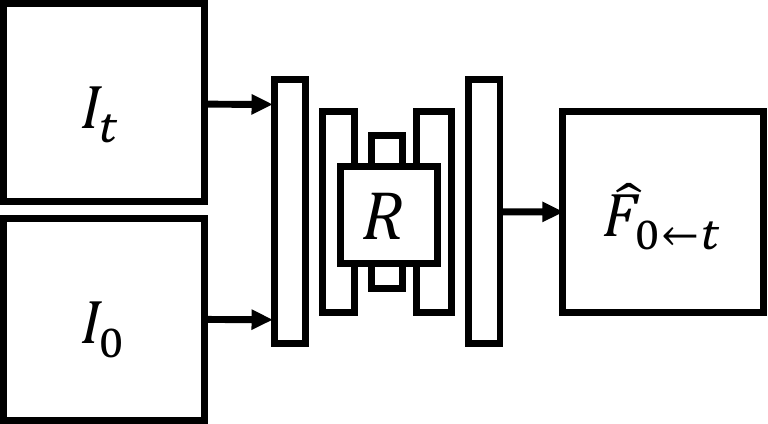}\vspace{2.5mm}
        \includegraphics[width=\textwidth]{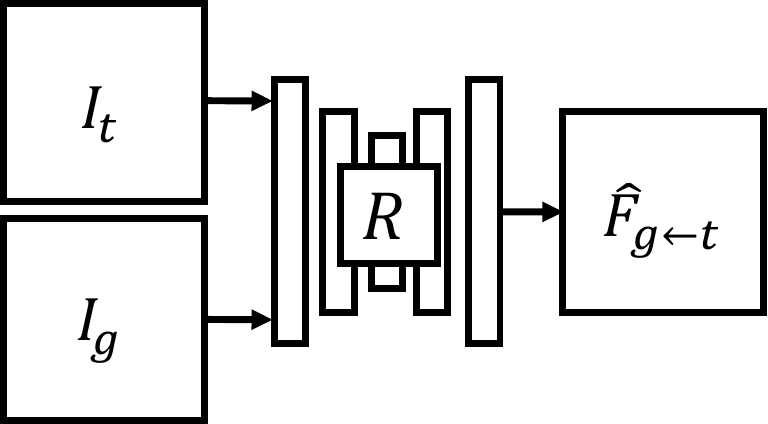}
        \caption{\small{Testing usage.}}
    \end{subfigure}
    \quad \quad
    \begin{subfigure}[b]{0.55\textwidth}
        \centering
        \includegraphics[width=\textwidth,trim={0 3mm 0 3mm},clip]{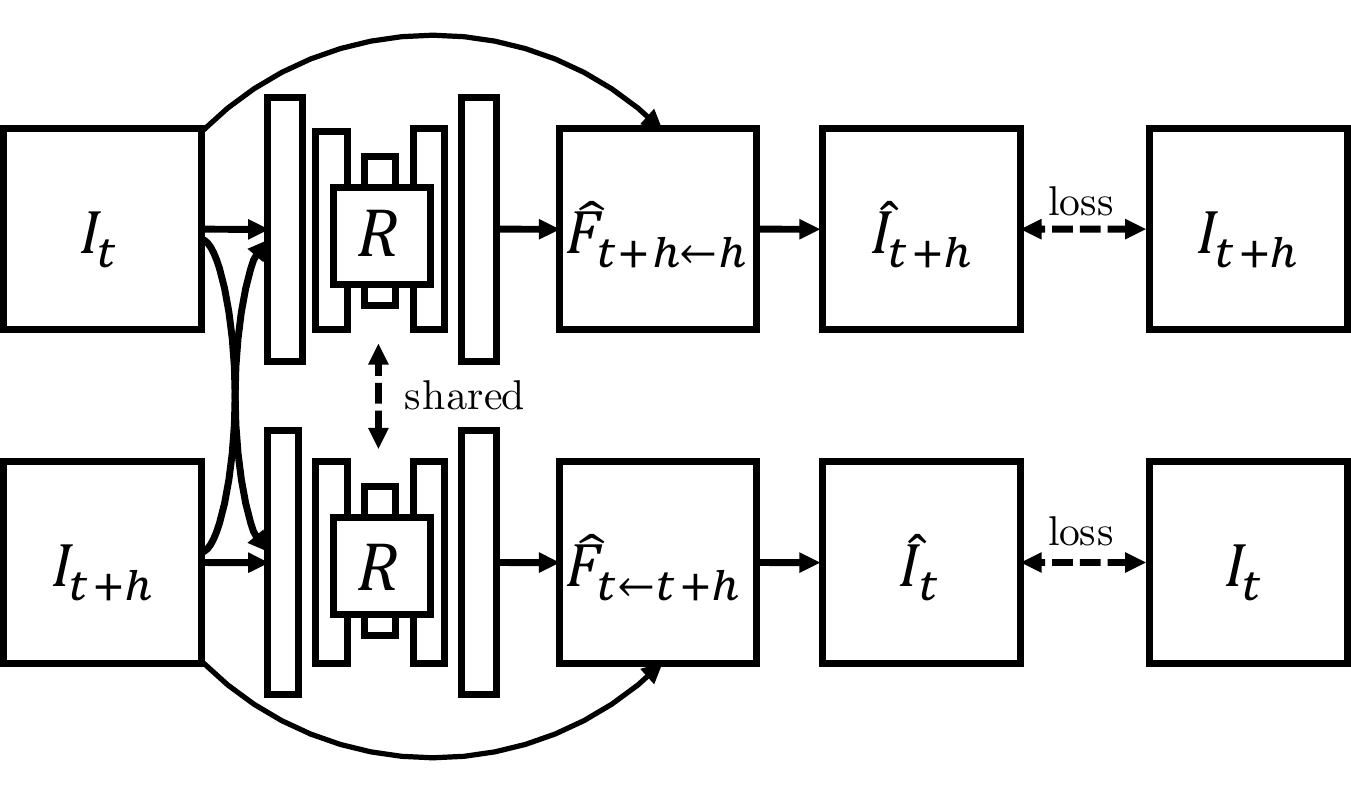}
        \caption{\small{Training usage.}}
        \label{fig:discrete}
    \end{subfigure}
    \vspace{-1mm}
    \caption{\small{(a) At test time the registration network registers the current image $I_t$ to the start image $I_0$ (top) and goal image $I_g$ (bottom), inferring the flow-fields $\hat{F}_{0 \leftarrow t}$ and $\hat{F}_{g \leftarrow t}$. (b) The registration network is trained by warping images from randomly selected timesteps along a trajectory to each other.
    }}
    \label{fig:registration_arch}
\end{figure}

 We will first describe the registration scheme at test time (see Figure~\ref{fig:registration_arch}(a)). We separately register the current image $I_t$ to the start image $I_0$ and to the goal image $I_g$ by passing it into the registration network $R$, implemented as a fully-convolutional neural network. Details about the archiecture are given in section \ref{subsec:training}. The registration network produces a flow map $\hat{F}_{0 \leftarrow t} \in \mathbb{R}^{H \times W \times 2}$, a vector field with the same size as the image, that describes the relative motion for every pixel between two frames:
\begin{align}
    \hat{F}_{0 \leftarrow t} = R(I_t, I_0) &&
    \hat{F}_{g \leftarrow t} = R(I_t, I_g)
\end{align}
The flow map $\hat{F}_{0 \leftarrow t}$ can be used to warp the image of the current time step $t$ to the start image $I_0$, and $\hat{F}_{g \leftarrow t}$ can be used to warp from $I_t$ to $I_g$ (see Figure \ref{fig:reg_single} for an illustration):
\begin{align}
    \hat{I}_0 = \hat{F}_{0 \leftarrow t} \diamond  I_t &&
    \hat{I}_g = \hat{F}_{g \leftarrow t} \diamond  I_t 
\end{align}
where $\diamond$ denotes a bilinear interpolation operator that interpolates the pixel value bilinearly with respect to a location $(x,y)$ and its four neighbouring pixels in the image. In essence for a current image $\hat{F}_{0 \leftarrow t}$ puts $I_t$ in correspondence with $I_0$, and $\hat{F}_{g \leftarrow t}$ puts $I_t$ in correspondence with $I_g$. As one might expect, warping works better for images that are closer to each other, and sometimes fails when the entities in the image are too far apart. The motivation for registering to both $I_0$ and $I_g$ is to increase accuracy and robustness. In principle, registering to either $I_0$ or $I_g$ is sufficient.

While the registration network is trained to perform a global registration between the images, we only evaluate it at the points $d_0$ and $d_g$ chosen by the user. This results in a cost function that ignores distractors. The flow map produced by the registration network is used to find the pixel locations corresponding to $d_0$ and $d_g$ in the current frame: 
\begin{align}
    \hat{d}_{0,t} = d_0 + \hat{F}_{0 \leftarrow t}(d_0) &&
    \hat{d}_{g,t} = d_g + \hat{F}_{g \leftarrow t}(d_g)
    \label{eqn:warped_pos}
\end{align}

\begin{figure}
    \centering
    \vspace{-0.1in}
    \includegraphics[width=1\textwidth]{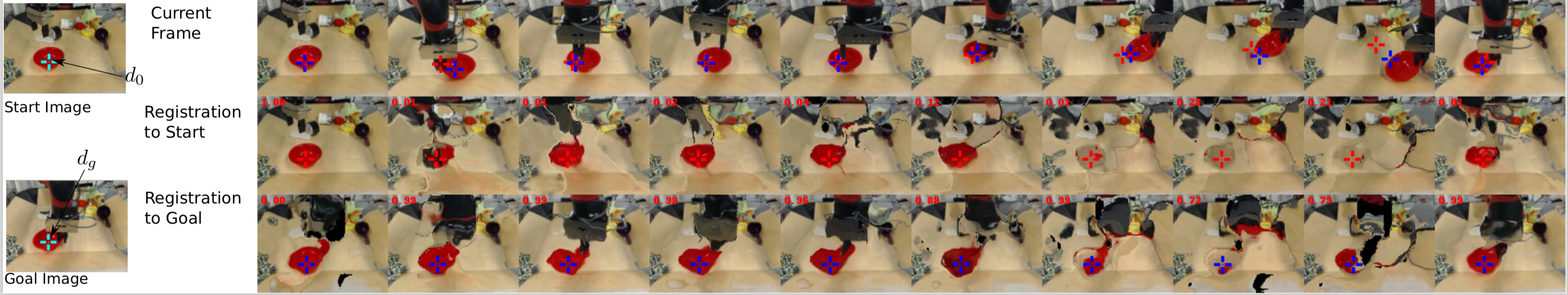}
    \caption{\small{Outputs of registration network. The first row shows the timesteps from left to right of a robot picking and moving a red bowl, the second row shows each image warped to the initial image via registration, and the third row shows the same for the goal image. A successful registration in this visualization would result in images that closely resemble the start- or goal image. In the first row, the locations where the designated pixel of the start image $d_0$ and the goal image $d_g$ are found are marked with red and blue crosses, respectively. It can be seen that the registration to the start image (red cross) is failing in the second to last time step, while the registration to the goal image (blue cross) succeeds for all time steps. The numbers in red, in the upper left corners indicate the trade off factors $\lambda$ between the views and are used as weighting factors for the planning cost. (Best viewed in PDF)}}
    \label{fig:tracking_overtime}
    \vspace{-0.2in}
\end{figure}

For simplicity, we describe the case with a single designated pixel. In practice, instead of a single flow vector $\hat{F}_{0 \leftarrow t}(d_0)$ and $\hat{F}_{g \leftarrow t}(d_g)$, we consider a neighborhood of flow-vectors around $d_0$ and $d_g$ and take the median in the $x$ and $y$ directions, making the registration more stable.
\autoref{fig:tracking_overtime} visualizes an example tracking result while the gripper is moving an object.

\vspace{-0.1cm}
\subsection{Planning Costs}
\label{subsec:planning_costs}
\vspace{-0.2cm}

Registration can fail when distances between objects in the images are large. During a trajectory, the registration to the first image typically becomes harder, while the registration to the goal image becomes easier. We propose a mechanism that estimates which image is registered correctly, allowing us to utilize only the successful registration for evaluating the planning cost. This mechanism gives a high weight $\lambda_i$ to pixel-distance costs $c_i$ associated with a designated pixel $\hat{d}_{i,t}$ that is tracked successfully and a low, ideally zero, weight to a designated pixel where the registration is poor. We propose to use the photometric distance between the true frame and the warped frame evaluated at $d_{0,i}$ and $d_{g,i}$ as an estimate for \emph{local} registration success. A low photometric error indicates that the registration network predicted a flow vector leading to a pixel with a similar color, thus indicating warping success. However this does not necessarily mean that the flow vector points to the correct location. For example, there could be several objects with the same color and the network could simply point to the wrong object. Letting $I_i(d_i)$ denote the pixel value in image $I_i$ for position $d_i$, and $\hat{I}_i(d_i)$ denote the corresponding pixel in the image warped by the registration function, we can define the general weighting factors $\lambda_i$ as
\begin{align}
\lambda_i =  \frac{||I_i(d_i) - \hat{I_i}(d_i)||_2^{-1}}{\sum^N_j ||I_j(d_j) - \hat{I}_j(d_j)||^{-1}_2}.
\label{eqn:cost_avg}
\end{align}
where $\hat{I}_i = \hat{F}_{i \leftarrow t} \diamond I_t$. The MPC cost is computed as the average of the costs $c_i$ weighted by $\lambda_i$, where each $c_i$ is the expected distance (see \autoref{eq:cost}) between the registered point $\hat{d}_{i,t}$ and the goal point $d_{g,i}$. Hence, the cost used for planning is $c = \sum_i \lambda_i c_i$.  In the case of the single view model and a single designated pixel, the index $i$ iterates over the start and goal image (and $N=2$).

The proposed weighting scheme can also be used with multiple designated pixels, as used in multi-task settings and multi-view models, which are explained in section \ref{sec:scalingup}. The index $i$ then also loops over the views and indices of the designated pixels.

\subsection{Training procedure}
\label{subsec:training}
For registration we use a deep convolutional neural network $R$ which takes in a pair of images and finds correspondences by warping one image to the other. The network is trained on the same data as the video-prediction model, but it does not share parameters with it.\footnote{in principle sharing parameters with the video-prediction model might be beneficial, however this is left for future work} Our approach is similar to the optic flow method proposed by \citet{meister2017unflow}. However, unlike this prior work, our method computes registrations for frames that might be many time steps apart, and the goal is not to extract optic flow, but rather to determine correspondences between potentially distant images. For training, two images are sampled at random times steps $t$ and $t+h$ along the trajectory and the images are warped to each other in both directions. 
\begin{align}
     \hat{I}_{t} = \hat{F}_{t \leftarrow t +h} \diamond  I_{t+h} &&
     \hat{I}_{t+h} = \hat{F}_{t+h \leftarrow t} \diamond  I_{t}
\end{align}
The network, which outputs $\hat{F}_{t \leftarrow t +h}$ and $\hat{F}_{t+h \leftarrow t}$ (see Figure~\ref{fig:registration_arch}), is trained to minimize the photometric distance between $\hat{I}_t$ and $I_t$ and $\hat{I}_{t+h}$ and $I_{t+h}$, in addition to a smoothness regularizer that penalizes abrupt changes in the outputted flow-field. The details of this loss function follow prior work \cite{meister2017unflow}. We found that gradually increasing the temporal distance $h$ between the images during training yielded better final accuracy, as it creates a learning curriculum. The temporal distance is linearly increased from 1 step to 8 steps at 20k SGD steps. In total 60k iterations were taken.

The network $R$ is implemented as a fully convolutional network taking in two images stacked together along the channel dimension. We use three convolutional layers each followed by a bilinear downsampling operation. This is passed into three layers of convolution each followed by a bilinear upsampling operation (all convolutions use stride 1). By using bilinear sampling for increasing or decreasing image sizes we avoid artifacts that are caused by strided convolutions and deconvolutions.

\section{Scaling up Visual Model-Predictive Control}
\label{sec:scalingup}
\paragraph{Extension to multiple cameras.}
Prior work has only considered visual MPC with a single camera~\cite{foresight,sna}, where objects are manipulated on a plane. To define goals in 3D, we extend visual MPC to include multiple camera views. Since tasks are defined in terms of pixel motion in 2D image space, the combination of multiple 2D tasks with cameras oriented appropriately defines a 3D task. In our experiments, we show that we can define 3D manipulation tasks, such as lifting an object from the table, that would be ambiguous using only a single camera view. The registration method described in the previous section is used separately per view to allow for dynamic retrying and solving temporally extended tasks. The planning costs from each view are combined using weighted averaging where the weights are provided by the registration network (see equation \ref{eqn:cost_avg}). 

\begin{figure}
	\vspace{-0.1in}
	\centering
	\includegraphics[width=1.0\textwidth]{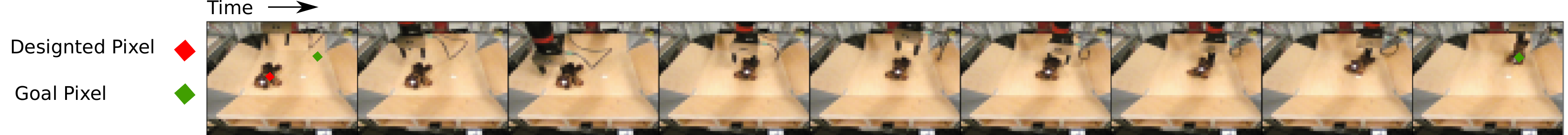}
	\caption{\small{Retrying behaviour of our method combining prehensile and non-prehensile manipulation. In the first 4 time instants shown the agent pushes the object. It then loses the object, and decides to grasp it pulling it all the way to the goal. Retrying is enabled by applying the learned registration to both camera views (here we only show the front view).}}
	\label{fig:push_grasp}
	\vspace{-0.2in}
\end{figure}

\vspace{-0.1in}
\paragraph{Combined prehensile and non-prehensile manipulation.}
In prior work on video-prediction based robotic manipulation \cite{sna, foresight}, the capabilities that emerged out of self-supervised learning were generally restricted to pushing and dragging objects. To enable more complex tasks, we also explore how visual MPC can enable behaviors that include picking and lifting objects for rearrangement. One of the main challenges with this is that random exploration is unlikely to pick up objects a sufficiently large fraction of the time to allow the model to learn grasping skills. To alleviate this challenge, we incorporate a simple ``reflex'' during data collection, where the gripper automatically closes when the height of the wrist above the table is lower than a small threshold. This reflex is inspired by the palmar reflex observed in infants~\cite{grasping_fetal}. With this primitive, about 20\% of training trajectories included some sort of grasp on an object. It is worth noting that, other than this reflex, no grasping-specific engineering was applied to the policy allowing a joint pushing and grasping policy to emerge, see figure \ref{fig:push_grasp}. In our experiments, we evaluate our method using data obtained both with and without the grasping reflex, evaluating both purely non-prehensile and combined prehensile and non-prehensile manipulation.

\vspace{-0.1cm}
\section{Experiments}
\vspace{-0.2cm}

\begin{wrapfigure}{r}{.35\columnwidth}
\vspace{-0.3in}
\centering
\includegraphics[width=0.35\columnwidth]{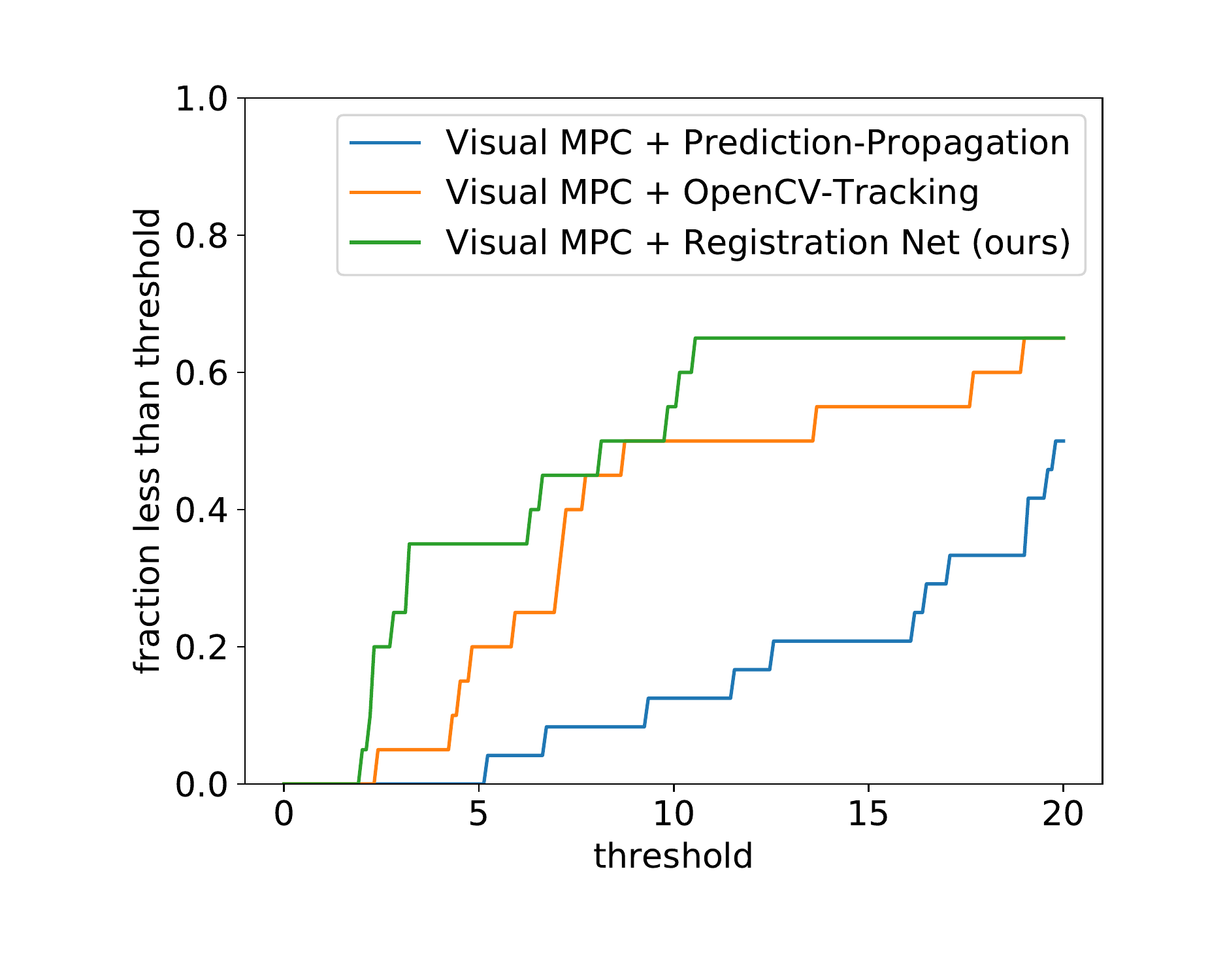}
\vspace{-0.9cm}
\caption{\small{Results for long pushing tasks with 20 objects not seen during training, showing fraction of runs where final distance is lower than threshold. Our method shows a clear gains over OpenCV tracking and predictor propagation.}}
\label{fig:push_bench_long}
\vspace{-0.1in}
\end{wrapfigure}

Videos and visualizations can be found on our supplementary webpage: \url{https://sites.google.com/view/robustness-via-retrying}.
Our experimental evaluation, conducted using two Sawyer robotic arms, evaluates the ability of our method to learn both prehensile and non-prehensile object relocation tasks entirely through autonomously collected data and self-supervision. In particular, we aim to answer the following questions: (1) How does our MPC approach with self-supervised image registration compare to alternative cost functions, such as off-the-shelf tracking and forward prediction via flow-based models? (2) When the robot can continuously retry a task using goal image registration, how much is the success rate improved for object relocation tasks? (3) Can we learn predictive models that enable both non-prehensile and prehensile object manipulation? In the appendix we also present additional experimental comparisons in a simulated environment.

\begin{wrapfigure}{r}{.35\columnwidth}
	\vspace{-0.2in}
	\centering
	\includegraphics[width=0.35\columnwidth]{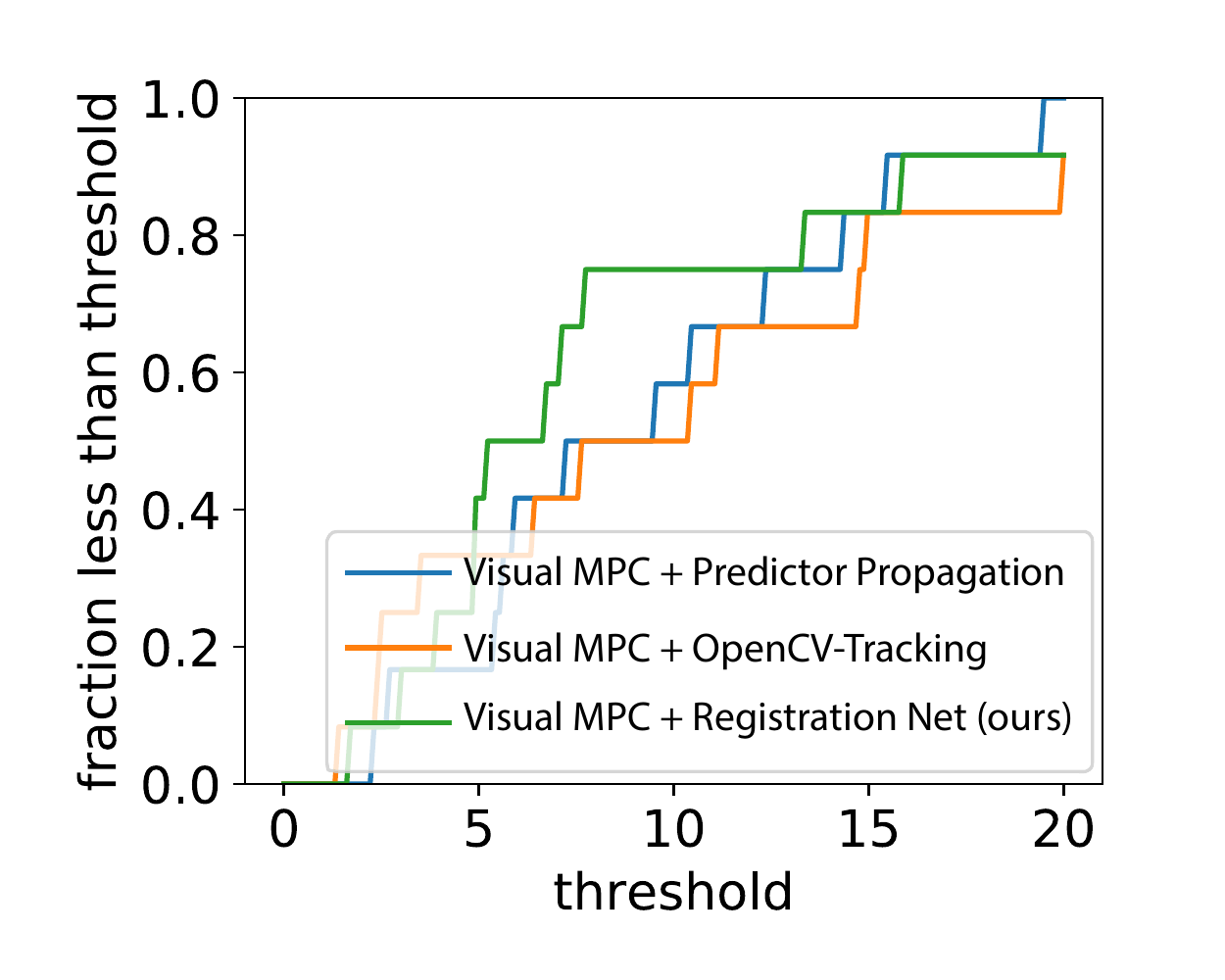}
	\vspace{-0.8cm}
	\caption{\small{Results for short pushing tasks.  Fraction of runs where final distance is lower than threshold.}}
	\label{fig:push_bench_short}
	\vspace{-0.2in}
\end{wrapfigure}


To train both our prediction and registration models, we collected 20,000 trajectories of pushing motions and 15,000 trajectories with gripper control, where the robot was allowed to randomly move and pick up objects (we use 150 objects for training and 5 objects for testing; see the appendix for details). The data collection process is fully autonomous, requiring human intervention only to replace and change out the objects in front of the robot.
The action space consists of Cartesian movements along the $x$, $y$, and $z$ axes, and changes in the azimuth orientation of the gripper, while the grasping action is triggered by a primitive as specified in Section~\ref{sec:scalingup}. For evaluation, we select novel objects that were never seen during training. The evaluation tasks required the robot to move objects in its environment from a starting state to a goal configuration, and performance is evaluated by measuring the distance between the final object position and goal position. In all experiments, the maximum episode length is 50 time steps.

\vspace{-0.1in}
\paragraph{Pushing with retrying.}
\begin{figure}
    \centering
    \includegraphics[width=1.0\textwidth]{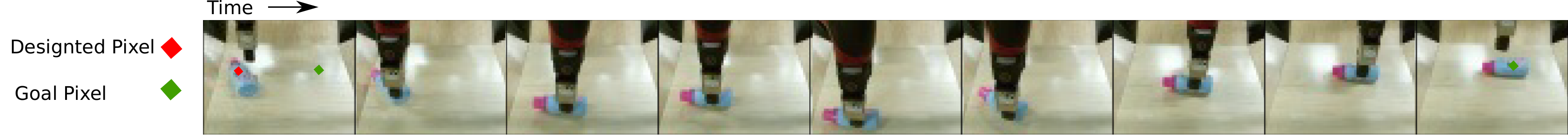}
    \caption{\small{Applying our method to a pushing task. The goal is to push the bottle to the green point. In the first 3 time instants the object behaves unexpectedly, moving down. The tracking then allows the robot to retry, allowing it to eventually bring the object to the goal.}}
    \label{fig:push_retry}
\end{figure}

In the first experiment, we aim to evaluate the performance of different visual MPC cost functions, including our proposed self-supervised registration cost. For this experiment, we disable the gripper control, requiring the robot to push objects to the target. 

We evaluate our method on 20 long-distance and 15 short-distance pushing tasks. For comparisons, we include an ablation where visual-MPC is used with a cost function that computes object locations using the ``multiple instance learning tracker'' (MIL) \cite{babenko2009visual} in OpenCV. This method represents a strong baseline trained with additional human supervision, which our method does not have access to.
We also compare to the visual MPC method proposed by \citet{sna},
which does not track the object explicitly, but relies on the flow-based video prediction model to keep track of the designated pixel, which we call ``predictor propagation.'' 

\autoref{fig:push_bench_long} illustrates that, on the long-distance benchmark, where the average distance between the object and its goal position is $30$ cm,
our approach not only outperforms prior work \cite{sna}, but also outperforms the hand-designed, supervised object tracker \cite{babenko2009visual}. Using our learned registration, the robot is more frequently able to successfully recover after mispredictions or occlusions (see \autoref{fig:push_retry}). In contrast, in the short distance benchmark, where the average push distance is $15$ cm
all methods perform comparably, as shown in \autoref{fig:push_bench_short}. These results indicate the importance of closed loop control in long-horizon tasks.

\begin{table}
	{\footnotesize
		\begin{center}
			\begin{tabular}{lcc}
				\toprule
				& Short & Long \\
				\midrule
			   Visual MPC $+$ predictor propagation  & 83\% & 20\% \\
  			   Visual MPC $+$ OpenCV tracking  & 83\%  & 45\% \\
			   Visual MPC $+$ registration network (Ours)  & 83\% & \textbf{66\%}  \\
				\bottomrule
			\end{tabular}
		\end{center}
	}
	\caption{\footnotesize Success rate for long-distance pushing benchmark with 20 different object/goal configurations and short-distance benchmark with 15 object/goal configurations. Success is defined as bringing the object closer than 15 pixels to the goal, where the complete image has size 48x64.}
	\label{table:res_longd}
	\vspace{-0.35in}
\end{table}



\vspace{-0.1in}
\paragraph{Combined prehensile and non-prehensile manipulation.}

\begin{wrapfigure}{r}{.35\columnwidth}
\vspace{-0.3in}
\centering
\hspace{-0.57cm}\includegraphics[width=0.39\columnwidth]{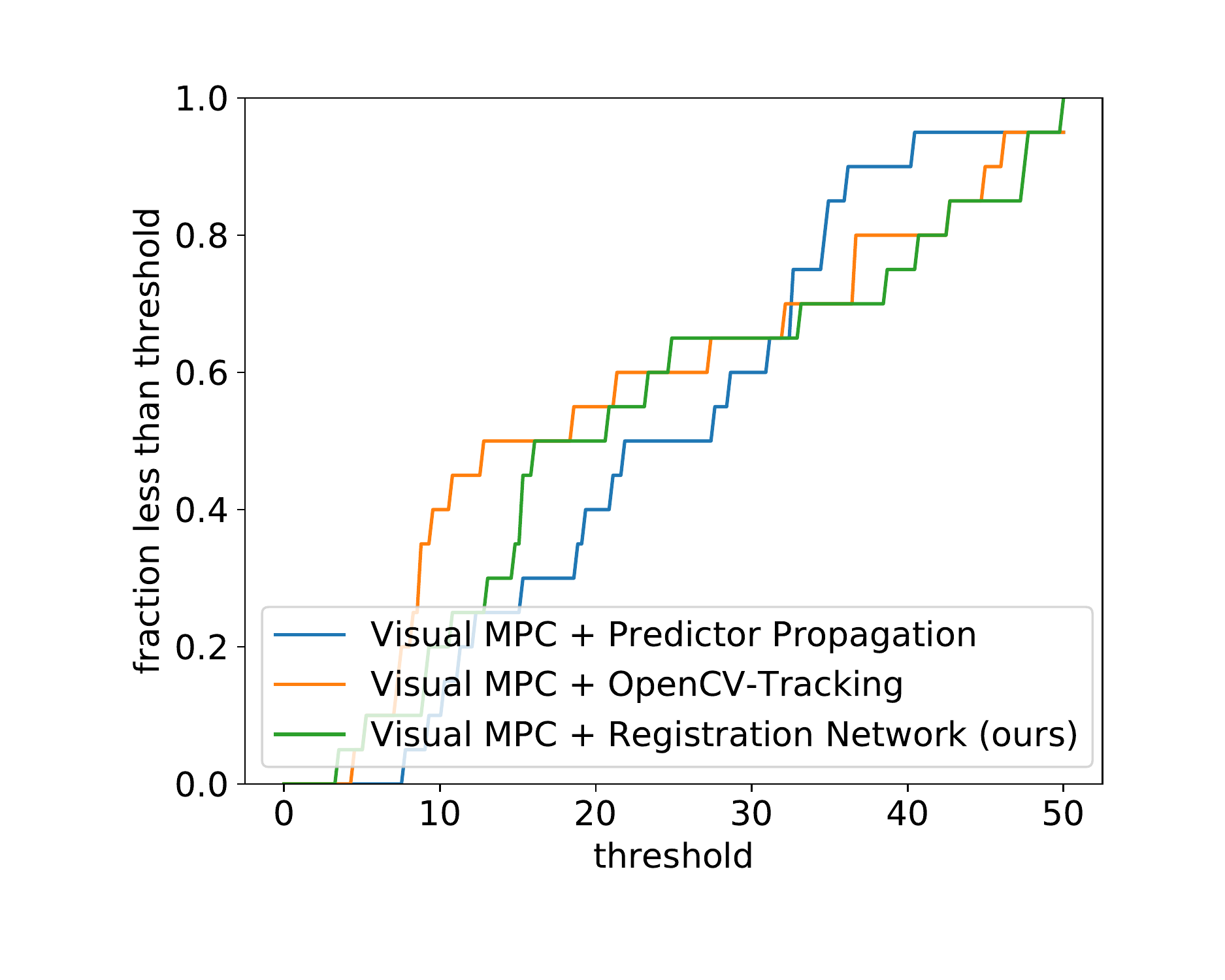}
\vspace{-0.9cm}
\caption{\small Results with combined prehensile and non-prehensile manipulation.}
\label{fig:grasp_bench}
\vspace{-0.3in}
\end{wrapfigure}

In the setting where the grasping reflex is enabled, the robot needs to decide whether to solve a task by grasping or pushing. We again use 20 object relocation tasks to measure the performance of each method, in terms of the final distance between the object and its goal location. We observe that, in the majority of the cases, the robot decides to grasp the object, as can be seen in the supplementary video. \autoref{fig:grasp_bench} shows the results of a benchmark on long-distance relocation tasks, which show that visual MPC combined with our registration method is comparable with the performance of visual MPC with the hand-designed and supervised OpenCV tracker, without requiring any human supervision.

\section{Discussion}

We demonstrate that using a cost function derived from learned image-to-image registration substantially improves performance on temporally extended object manipulation tasks by closing the control loop. Our experiments show a large improvement in success rate compared to a prior open-loop visual MPC method~\cite{sna}. We further show that, by including a simple grasping ``reflex'' inspired by the palmar reflex in infants, we can efficiently learn both non-prehensile and prehensile object relocation skills in a purely self-supervised way, allowing the robot to plan to pick up and move objects when necessary.  
Interesting challenges for future work are multi-object relocation tasks, which could be addressed by adding an abstract long-term planner on top of the presented framework.





\bibliography{bib}  

\begin{thebibliography}{44}
\providecommand{\natexlab}[1]{#1}
\providecommand{\url}[1]{\texttt{#1}}
\expandafter\ifx\csname urlstyle\endcsname\relax
  \providecommand{\doi}[1]{doi: #1}\else
  \providecommand{\doi}{doi: \begingroup \urlstyle{rm}\Url}\fi

\bibitem[Pinto and Gupta(2016)]{lerrel}
L.~Pinto and A.~Gupta.
\newblock Supersizing self-supervision: Learning to grasp from 50k tries and
  700 robot hours.
\newblock In \emph{International Conference on Robotics and Automation (ICRA)},
  2016.

\bibitem[Levine et~al.(2016)Levine, Pastor, Krizhevsky, Ibarz, and
  Quillen]{google_handeye}
S.~Levine, P.~Pastor, A.~Krizhevsky, J.~Ibarz, and D.~Quillen.
\newblock Learning hand-eye coordination for robotic grasping with deep
  learning and large-scale data collection.
\newblock \emph{International Journal of Robotics Research (IJRR)}, 2016.

\bibitem[Zeng et~al.(2018)Zeng, Song, Welker, Lee, Rodriguez, and
  Funkhouser]{princeton_pushgrasp}
A.~Zeng, S.~Song, S.~Welker, J.~Lee, A.~Rodriguez, and T.~Funkhouser.
\newblock Learning synergies between pushing and grasping with self-supervised
  deep reinforcement learning.
\newblock \emph{arXiv preprint arXiv:1803.09956}, 2018.

\bibitem[Finn and Levine(2017)]{foresight}
C.~Finn and S.~Levine.
\newblock Deep visual foresight for planning robot motion.
\newblock In \emph{International Conference on Robotics and Automation (ICRA)},
  2017.

\bibitem[Ebert et~al.(2017)Ebert, Finn, Lee, and Levine]{sna}
F.~Ebert, C.~Finn, A.~X. Lee, and S.~Levine.
\newblock Self-supervised visual planning with temporal skip connections.
\newblock \emph{CoRR}, abs/1710.05268, 2017.
\newblock URL \url{http://arxiv.org/abs/1710.05268}.

\bibitem[Agrawal et~al.(2016)Agrawal, Nair, Abbeel, Malik, and Levine]{pulkit}
P.~Agrawal, A.~V. Nair, P.~Abbeel, J.~Malik, and S.~Levine.
\newblock Learning to poke by poking: Experiential learning of intuitive
  physics.
\newblock In \emph{Advances in Neural Information Processing Systems}, 2016.

\bibitem[Byravan et~al.(2017)Byravan, Leeb, Meier, and Fox]{se3_control}
A.~Byravan, F.~Leeb, F.~Meier, and D.~Fox.
\newblock {SE3-Pose-Nets}: Structured deep dynamics models for visuomotor
  planning and control.
\newblock \emph{arXiv preprint arXiv:1710.00489}, 2017.

\bibitem[Gandhi et~al.(2017)Gandhi, Pinto, and Gupta]{crashing}
D.~Gandhi, L.~Pinto, and A.~Gupta.
\newblock Learning to fly by crashing.
\newblock \emph{arXiv preprint arXiv:1704.05588}, 2017.

\bibitem[Kahn et~al.(2017)Kahn, Villaflor, Pong, Abbeel, and
  Levine]{greg_kahn_uncertainty}
G.~Kahn, A.~Villaflor, V.~Pong, P.~Abbeel, and S.~Levine.
\newblock Uncertainty-aware reinforcement learning for collision avoidance.
\newblock \emph{arXiv preprint arXiv:1702.01182}, 2017.

\bibitem[Jagersand and Nelson(1995)]{jagersand1995visual}
M.~Jagersand and R.~Nelson.
\newblock Visual space task specification, planning and control.
\newblock In \emph{International Symposium on Computer Vision}, 1995.

\bibitem[Deguchi and Takahashi(1999)]{deguchi1999image}
K.~Deguchi and I.~Takahashi.
\newblock Image-based simultaneous control of robot and target object motions
  by direct-image-interpretation method.
\newblock In \emph{International Conference on Intelligent Robots and Systems
  (IROS)}, 1999.

\bibitem[Watter et~al.(2015)Watter, Springenberg, Boedecker, and
  Riedmiller]{e2c}
M.~Watter, J.~Springenberg, J.~Boedecker, and M.~Riedmiller.
\newblock Embed to control: A locally linear latent dynamics model for control
  from raw images.
\newblock In \emph{Advances in neural information processing systems}, 2015.

\bibitem[Finn et~al.(2016)Finn, Tan, Duan, Darrell, Levine, and Abbeel]{dsae}
C.~Finn, X.~Y. Tan, Y.~Duan, T.~Darrell, S.~Levine, and P.~Abbeel.
\newblock Deep spatial autoencoders for visuomotor learning.
\newblock In \emph{Robotics and Automation (ICRA), 2016 IEEE International
  Conference on}. IEEE, 2016.

\bibitem[Hutchinson et~al.(1996)Hutchinson, Hager, and
  Corke]{hutchinson1996tutorial}
S.~Hutchinson, G.~D. Hager, and P.~I. Corke.
\newblock A tutorial on visual servo control.
\newblock \emph{IEEE transactions on robotics and automation}, 12\penalty0
  (5):\penalty0 651--670, 1996.

\bibitem[Kragic and Christensen(2002)]{kragic2002survey}
D.~Kragic and H.~I. Christensen.
\newblock Survey on visual servoing for manipulation.
\newblock \emph{Computational Vision and Active Perception Laboratory,
  Fiskartorpsv}, 15, 2002.

\bibitem[DeSouza and Kak(2002)]{desouza2002survey}
G.~N. DeSouza and A.~C. Kak.
\newblock Vision for mobile robot navigation: A survey.
\newblock \emph{IEEE transactions on pattern analysis and machine
  intelligence}, 24\penalty0 (2):\penalty0 237--267, 2002.

\bibitem[Feddema and Mitchell(1989)]{feddema1989vision}
J.~T. Feddema and O.~R. Mitchell.
\newblock Vision-guided servoing with feature-based trajectory generation (for
  robots).
\newblock \emph{IEEE Transactions on Robotics and Automation}, 5\penalty0
  (5):\penalty0 691--700, 1989.

\bibitem[Espiau et~al.(1992)Espiau, Chaumette, and Rives]{espiau1992servo}
B.~Espiau, F.~Chaumette, and P.~Rives.
\newblock A new approach to visual servoing in robotics.
\newblock \emph{IEEE Transactions on Robotics and Automation}, 8\penalty0
  (3):\penalty0 313--326, 1992.

\bibitem[Wilson et~al.(1996)Wilson, Williams~Hulls, and
  Bell]{wilson1996relative}
W.~J. Wilson, C.~C. Williams~Hulls, and G.~S. Bell.
\newblock Relative end-effector control using cartesian position based visual
  servoing.
\newblock \emph{IEEE Transactions on Robotics and Automation}, 12\penalty0
  (5):\penalty0 684--696, 1996.

\bibitem[Caron et~al.(2013)Caron, Marchand, and Mouaddib]{caron2013photometric}
G.~Caron, E.~Marchand, and E.~M. Mouaddib.
\newblock Photometric visual servoing for omnidirectional cameras.
\newblock \emph{Autonomous Robots}, 35\penalty0 (2-3):\penalty0 177--193, 2013.

\bibitem[Saxena et~al.(2017)Saxena, Pandya, Kumar, Gaud, and
  Krishna]{saxena2017servoing}
A.~Saxena, H.~Pandya, G.~Kumar, A.~Gaud, and K.~M. Krishna.
\newblock Exploring convolutional networks for end-to-end visual servoing.
\newblock In \emph{International Conference on Robotics and Automation (ICRA)},
  2017.

\bibitem[Bateux et~al.(2018)Bateux, Marchand, Leitner, and
  Chaumette]{bateux2018servoing}
Q.~Bateux, E.~Marchand, J.~Leitner, and F.~Chaumette.
\newblock Visual servoing from deep neural networks.
\newblock In \emph{International Conference on Robotics and Automation (ICRA)},
  2018.

\bibitem[Lee et~al.(2017)Lee, Levine, and Abbeel]{lee2017servoing}
A.~X. Lee, S.~Levine, and P.~Abbeel.
\newblock Learning visual servoing with deep features and fitted {Q}-iteration.
\newblock \emph{International Conference on Learning Representations (ICLR)},
  2017.

\bibitem[Camacho and Alba(2013)]{camacho2013model}
E.~F. Camacho and C.~B. Alba.
\newblock \emph{Model predictive control}.
\newblock Springer Science \& Business Media, 2013.

\bibitem[Shim et~al.(2003)Shim, Kim, and Sastry]{shim2003decentralized}
D.~H. Shim, H.~J. Kim, and S.~Sastry.
\newblock Decentralized nonlinear model predictive control of multiple flying
  robots.
\newblock In \emph{Decision and control, 2003. Proceedings. 42nd IEEE
  conference on}, volume~4, pages 3621--3626. IEEE, 2003.

\bibitem[Allibert et~al.(2010)Allibert, Courtial, and
  Chaumette]{allibert2010predictive}
G.~Allibert, E.~Courtial, and F.~Chaumette.
\newblock Predictive control for constrained image-based visual servoing.
\newblock \emph{IEEE Transactions on Robotics}, 26\penalty0 (5):\penalty0
  933--939, 2010.

\bibitem[Howard et~al.(2010)Howard, Green, and Kelly]{howard2010receding}
T.~M. Howard, C.~J. Green, and A.~Kelly.
\newblock Receding horizon model-predictive control for mobile robot navigation
  of intricate paths.
\newblock In \emph{Field and Service Robotics}, pages 69--78. Springer, 2010.

\bibitem[Williams et~al.(2017)Williams, Wagener, Goldfain, Drews, Rehg, Boots,
  and Theodorou]{williams2017information}
G.~Williams, N.~Wagener, B.~Goldfain, P.~Drews, J.~M. Rehg, B.~Boots, and E.~A.
  Theodorou.
\newblock Information theoretic mpc for model-based reinforcement learning.
\newblock In \emph{Robotics and Automation (ICRA), 2017 IEEE International
  Conference on}, pages 1714--1721. IEEE, 2017.

\bibitem[Lenz and Saxena(2015)]{deep_mpc}
I.~Lenz and A.~Saxena.
\newblock Deepmpc: Learning deep latent features for model predictive control.
\newblock In \emph{In RSS}. Citeseer, 2015.

\bibitem[Lucas et~al.(1981)Lucas, Kanade, et~al.]{lucas1981iterative}
B.~D. Lucas, T.~Kanade, et~al.
\newblock An iterative image registration technique with an application to
  stereo vision.
\newblock 1981.

\bibitem[Brox et~al.(2004)Brox, Bruhn, Papenberg, and Weickert]{brox2004high}
T.~Brox, A.~Bruhn, N.~Papenberg, and J.~Weickert.
\newblock High accuracy optical flow estimation based on a theory for warping.
\newblock In \emph{European conference on computer vision}. Springer, 2004.

\bibitem[Babenko et~al.(2009)Babenko, Yang, and Belongie]{babenko2009visual}
B.~Babenko, M.-H. Yang, and S.~Belongie.
\newblock Visual tracking with online multiple instance learning.
\newblock In \emph{Computer Vision and Pattern Recognition, 2009. CVPR 2009.
  IEEE Conference on}, pages 983--990. IEEE, 2009.

\bibitem[Mei and Ling(2009)]{mei2009robust}
X.~Mei and H.~Ling.
\newblock Robust visual tracking using l 1 minimization.
\newblock In \emph{Computer Vision, 2009 IEEE 12th International Conference
  on}, pages 1436--1443. IEEE, 2009.

\bibitem[Meister et~al.(2017)Meister, Hur, and Roth]{meister2017unflow}
S.~Meister, J.~Hur, and S.~Roth.
\newblock Unflow: Unsupervised learning of optical flow with a bidirectional
  census loss.
\newblock \emph{arXiv preprint arXiv:1711.07837}, 2017.

\bibitem[Lee et~al.(2018)Lee, Zhang, Ebert, Abbeel, Finn, and Levine]{savp}
A.~X. Lee, R.~Zhang, F.~Ebert, P.~Abbeel, C.~Finn, and S.~Levine.
\newblock Stochastic adversarial video prediction.
\newblock \emph{arXiv preprint arXiv:1804.01523}, 2018.

\bibitem[Rubinstein and Kroese(2013)]{cem-rk-13}
R.~Y. Rubinstein and D.~P. Kroese.
\newblock \emph{The cross-entropy method: a unified approach to combinatorial
  optimization, Monte-Carlo simulation and machine learning}.
\newblock Springer Science \& Business Media, 2013.

\bibitem[Sherer(1993)]{grasping_fetal}
D.~Sherer.
\newblock Fetal grasping at 16 weeks' gestation.
\newblock \emph{Journal of ultrasound in medicine}, 12\penalty0 (6), 1993.

\bibitem[Hermans et~al.(2013)Hermans, Li, Rehg, and
  Bobick]{hermans2013learning}
T.~Hermans, F.~Li, J.~M. Rehg, and A.~F. Bobick.
\newblock Learning contact locations for pushing and orienting unknown objects.
\newblock Georgia Institute of Technology, 2013.

\bibitem[Salganicoff et~al.(1993)Salganicoff, Metta, Oddera, and
  Sandini]{salganicoff1993vision}
M.~Salganicoff, G.~Metta, A.~Oddera, and G.~Sandini.
\newblock \emph{A vision-based learning method for pushing manipulation},
  volume~54.
\newblock University of Pennsylvania, 1993.

\bibitem[Goldfeder et~al.(2009)Goldfeder, Ciocarlie, Peretzman, Dang, and
  Allen]{goldfeder2009data}
C.~Goldfeder, M.~Ciocarlie, J.~Peretzman, H.~Dang, and P.~K. Allen.
\newblock Data-driven grasping with partial sensor data.
\newblock In \emph{Intelligent Robots and Systems, 2009. IROS 2009. IEEE/RSJ
  International Conference on}, pages 1278--1283. Citeseer, 2009.

\bibitem[Mahler et~al.(2017)Mahler, Liang, Niyaz, Laskey, Doan, Liu, Ojea, and
  Goldberg]{mahler2017dex}
J.~Mahler, J.~Liang, S.~Niyaz, M.~Laskey, R.~Doan, X.~Liu, J.~A. Ojea, and
  K.~Goldberg.
\newblock Dex-net 2.0: Deep learning to plan robust grasps with synthetic point
  clouds and analytic grasp metrics.
\newblock \emph{arXiv preprint arXiv:1703.09312}, 2017.

\bibitem[Lenz et~al.(2015)Lenz, Lee, and Saxena]{lenz2015deep}
I.~Lenz, H.~Lee, and A.~Saxena.
\newblock Deep learning for detecting robotic grasps.
\newblock \emph{The International Journal of Robotics Research}, 34\penalty0
  (4-5):\penalty0 705--724, 2015.

\bibitem[Zeng et~al.(2017)Zeng, Song, Yu, Donlon, Hogan, Bauza, Ma, Taylor,
  Liu, Romo, et~al.]{zeng2017robotic}
A.~Zeng, S.~Song, K.-T. Yu, E.~Donlon, F.~R. Hogan, M.~Bauza, D.~Ma, O.~Taylor,
  M.~Liu, E.~Romo, et~al.
\newblock Robotic pick-and-place of novel objects in clutter with
  multi-affordance grasping and cross-domain image matching.
\newblock \emph{arXiv preprint arXiv:1710.01330}, 2017.

\bibitem[Todorov et~al.(2012)Todorov, Erez, and Tassa]{todorov2012mujoco}
E.~Todorov, T.~Erez, and Y.~Tassa.
\newblock Mujoco: A physics engine for model-based control.
\newblock In \emph{Intelligent Robots and Systems (IROS), 2012 IEEE/RSJ
  International Conference on}, pages 5026--5033. IEEE, 2012.

\end{thebibliography}

\newpage
\appendix

{\footnotesize
\part*{Appendix}

\subsection*{Simulated Experiments}

\begin{wrapfigure}{r}{.34\columnwidth}
	\centering
	\includegraphics[width=0.3\columnwidth]{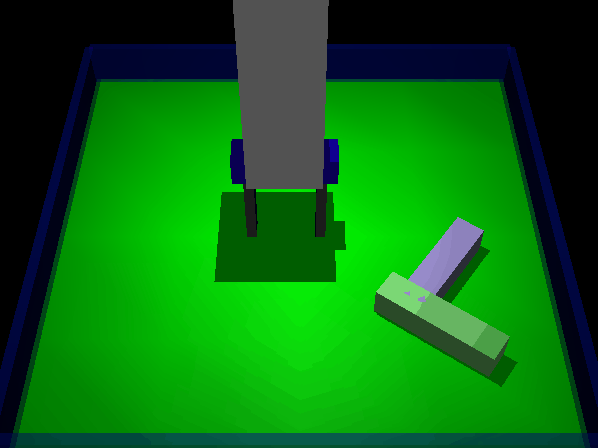}
	\caption{\small{Block pushing simulator}}
	\label{fig:sim}
\end{wrapfigure}

In order to provide a more controlled comparison, we also set up a realistic simulation environment using MuJoCo \cite{todorov2012mujoco}, which includes a robotic manipulator controlled via Cartesian position control, similar to our real world setup, pushing randomly-generated L-shaped objects with random colors (see details in supplementary materials). 
We trained the same video prediction model in this environment, and set up 50 evaluation tasks where blocks must be pushed to target locations with maximum episode lengths of 120 steps. 
\begin{wrapfigure}{r}{.4\columnwidth}
	\centering
	\includegraphics[width=0.4\columnwidth]{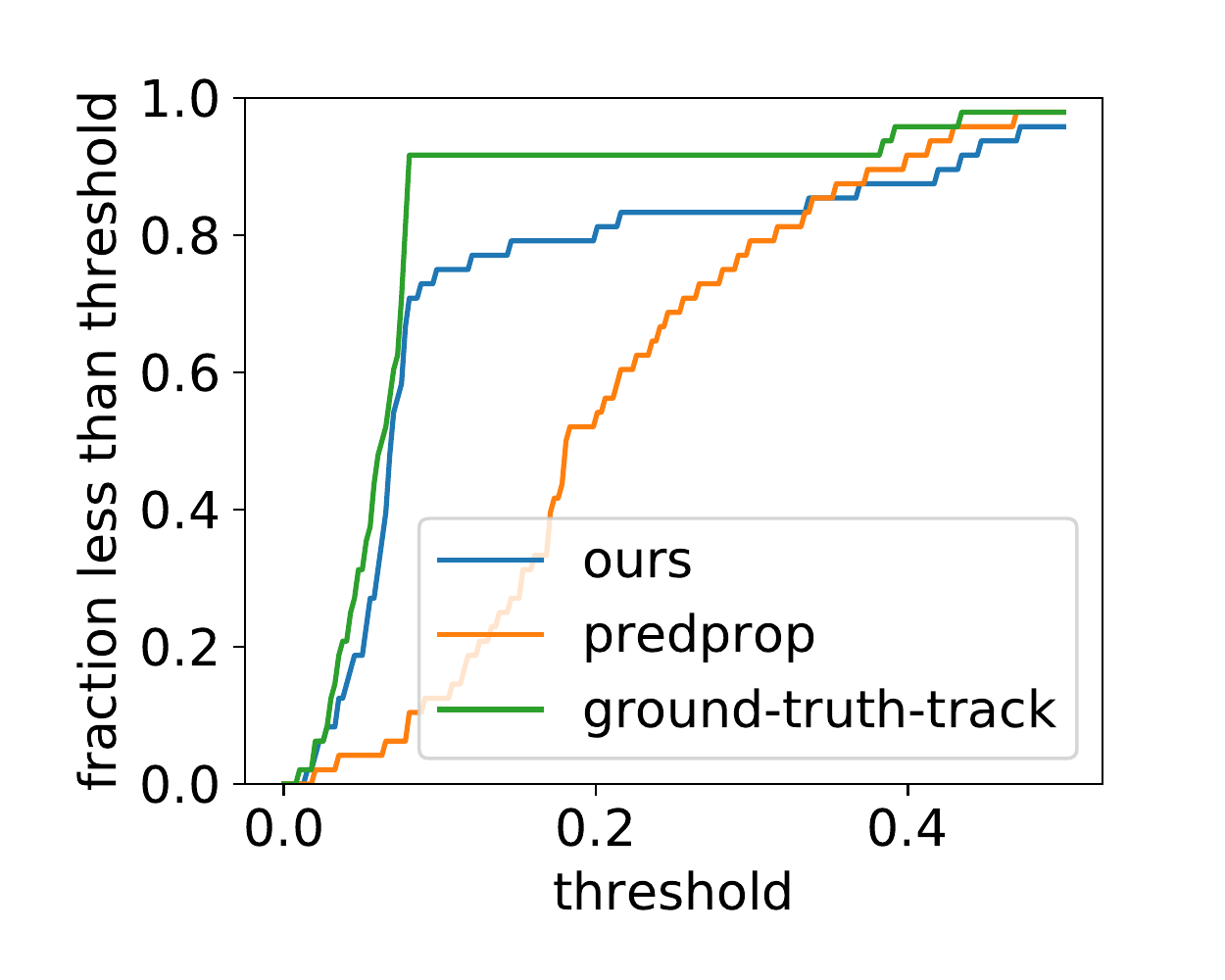}
	\caption{\small{Simulated evaluation. Fraction of trajectories with final object distance lower than threshold (higher is better).}}
	\label{fig:sim_bench}
\end{wrapfigure}

We  compare our proposed registration-based method, ``predictor propagation,'' and ground-truth registration obtained from the simulator, which provides an oracle upper bound on registration performance. \autoref{fig:sim_bench} shows the results of this simulated evaluation, where the x-axis shows different distance thresholds, and the y-axis shows the fraction of evaluation scenarios where each method pushed the object within that threshold. We can see that, for thresholds around 0.1, our method drastically outperforms predictor propagation (i.e., prior work~\cite{sna}), and has a relatively modest gap in performance against ground-truth tracking. This indicates that our registration method is highly effective in guiding visual MPC, despite being entirely self-supervised.

In our simulated experiments, we use end-effector position control with the arm illustrated in Figure~\ref{fig:sim}. In this environment, the video prediction model was trained with using 60,000 training trajectories.

\section*{Planner Implementation Details}

We use the cross-entropy method \cite{cem-rk-13} to optimize the action sequence with respect to the cost function. At the first iteration 400 samples and at later iterations 200 samples are taken and  a Gaussian is fitted to the best $5\%$ of the samples. We use 3 CEM iterations. The length of the prediction horizon is 15, to reduce the search space we use an action repeat of 3, so that only a sequence of 5 independent actions needs to be optimized. Since the action space is 4 (x,y,z,rotation) a total of 20 variables are optimized.

\section*{Improvements to online optimization procedure}
In the visual MPC setting the action sequences found by the optimizer can be very different between execution times steps (not to be confused with prediction time steps). For example at one time step the optimizer might find a pushing action leading towards the goal and in the next time step it determines a grasping action to the optimal to reach the goal. Naive replanning at every time step can result in alternating between a pushing attempt and a grasping attempt indefinitely causing the agent to get stuck and not making any progress towards to goal. 

We show that we can resolve this problem by modifying the sampling distribution of the first iteration of CEM so that the optimizer commits to the plan found in the previous time step. In prior work \cite{sna} the sampling distribution at first iteration of CEM is chosen to be a Gaussian with diagonal covariance matrix and zero mean. We instead use the best action sequence found in the optimization of the previous time step as the mean. Since this action sequence is optimized for the previous time step we only use the values $a_{1:T}$ and omit the first action, where $T$ is the prediction horizon. To sample actions close to the action sequence from the previous time step we reduce the entries of the diagonal covariance matrix for the first $T-1$ time steps. It is crucial that the last entry of the covariance matrix at the end of the horizon is not reduced otherwise no exploration could happen for the last time step causing poor performance at later time steps.
}

\end{document}